\pdfoutput=1
\documentclass[11pt]{article}
\usepackage{authblk}

\usepackage{ACL2023}
\usepackage{times}
\usepackage{latexsym}
\usepackage[T1]{fontenc}
\usepackage[utf8]{inputenc}
\usepackage{microtype}
\usepackage{inconsolata}
\usepackage{graphicx}
\usepackage{fvextra}
\usepackage{todonotes}
\usepackage{tikz}
\usepackage[subtle]{savetrees}

\usetikzlibrary{patterns.meta} 
\usepackage[most]{tcolorbox}

\DeclareUnicodeCharacter{26A0}{} 
\DeclareUnicodeCharacter{FE0F}{}                      
\DeclareUnicodeCharacter{2265}{$\geq$}  

\usepackage{fvextra}   
\usepackage{mdframed}  

\mdfdefinestyle{pa@box}{%
  linewidth=0.6pt,
  roundcorner=3pt,
  skipabove=\baselineskip,
  skipbelow=\baselineskip,
  innerleftmargin=8pt,
  innerrightmargin=8pt,
  innertopmargin=6pt,
  innerbottommargin=6pt
}

\newenvironment{promptanswer}{%
  \begin{mdframed}[style=pa@box]%
}{%
  \end{mdframed}%
}

\newenvironment{PAprompt}{%
  \par\noindent\textbf{Prompt}\par\vspace{0.3\baselineskip}%
  \VerbatimEnvironment
  \begin{Verbatim}[breaklines=true,breakanywhere=true,%
                   breaksymbolleft={},breaksymbolright={},breakindent=0pt,%
                   fontsize=\small]%
}{\end{Verbatim}}

\newenvironment{PAanswer}{%
  \par\medskip\noindent\textbf{Answer}\par\vspace{0.3\baselineskip}%
  \VerbatimEnvironment
  \begin{Verbatim}[breaklines=true,breakanywhere=true,%
                   breaksymbolleft={},breaksymbolright={},breakindent=0pt,%
                   fontsize=\small]%
}{\end{Verbatim}}

\newcommand{\easySymbol}{\tikz[baseline=-0.2ex]\draw (0,0) rectangle (0.8em,0.8em);}
\newcommand{\hardSymbol}{%
  \tikz[baseline=-0.2ex]
    \filldraw[
      pattern={Dots[distance=3pt, radius=0.4pt]},
      pattern color=black
    ] (0,0) rectangle (0.8em,0.8em);%
}

\title{Reasoning Core: A Scalable RL Environment for LLM Symbolic Reasoning 
}

\newlength{\myMheight}
\newcommand{\github}{\includegraphics[height=\myMheight]{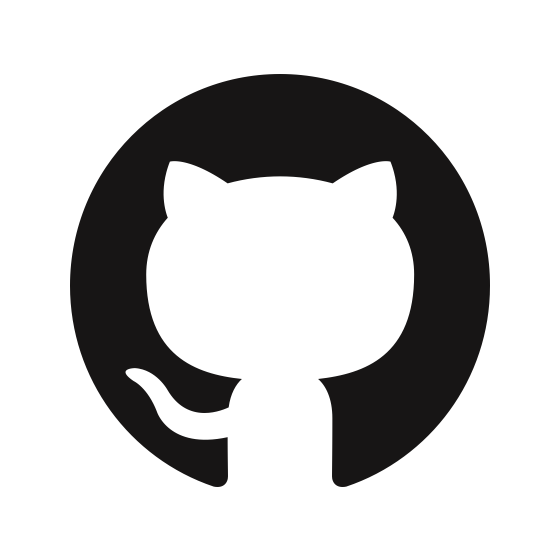}}
\newcommand{\hf}{\includegraphics[height=\myMheight]{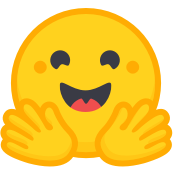}}
\newcommand{\pitl}{\includegraphics[height=\myMheight]{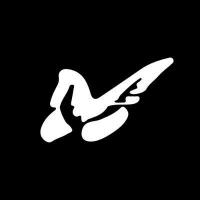}}

\author[]{Valentin Lacombe}
\author[]{Valentin Quesnel}
\author[]{Damien Sileo}
\affil[]{Univ. Lille, Inria, CNRS, Centrale Lille, UMR 9189 - CRIStAL, F-59000 Lille, France}
\affil[]{\url{damien.sileo@inria.fr}}

\begin{document}
\maketitle

\begin{abstract}
We introduce Reasoning Core, a new scalable environment for Reinforcement Learning with Verifiable Rewards (RLVR), designed to advance foundational symbolic reasoning in Large Language Models (LLMs). Unlike existing benchmarks that focus on games or isolated puzzles, Reasoning Core procedurally generates problems across core formal domains, including PDDL planning, first-order logic, context-free grammar parsing, causal reasoning, and system equation solving. The environment is built on key design principles of high-generality problem distributions, verification via external tools, and continuous difficulty control, which together provide a virtually infinite supply of novel training instances. Initial zero-shot evaluations with frontier LLMs confirm the difficulty of Reasoning Core's tasks, positioning it as a promising resource to improve the reasoning capabilities of future models.

\end{abstract}

\section{Introduction}

The reasoning capabilities of Large Language Models (LLMs) have advanced significantly, with state-of-the-art performance often attributed to Reinforcement Learning with Verifiable Rewards (RLVR)~\cite{r1, tulu3, o1}. This paradigm trains models using outcome-based feedback from environments where solution correctness is algorithmically verified, enabling models to learn complex, open-ended reasoning processes.

However, RLVR's efficacy hinges on the availability of high-quality, scalable training data. Current approaches frequently rely on fixed, human-curated benchmarks~\cite{cobbe2021training} or noisy, finite internet-scraped content~\cite{villalobos2022will}, creating a fundamental scalability bottleneck. Procedural generation has emerged as a promising direction to address this. Libraries like Reasoning Gym~\cite{stojanovski2025reasoning} represent a significant step, offering over 100 algorithmically verifiable environments with adjustable difficulty, spanning domains from games to arithmetic.

Despite this progress, many existing generative benchmarks such as  Reasoning Gym focus on tasks like puzzles, games, or templates yielding limited data diversity. While these are invaluable, they may not fully probe the foundational cognitive capacities required for general-purpose reasoning. They often test cleverness within constrained rule sets rather than the ability to handle the full, expressive generality of domains like formal logic, symbolic planning, or inductive reasoning.

To bridge this gap, we introduce \textbf{Reasoning Core}, a library of reasoning environments focused on fundamental symbolic domains. In contrast to broad-coverage libraries, Reasoning Core offers fewer, but more general and foundational tasks: PDDL planning in randomly generated domains, First-order logic with equality, Context-free grammar parsing, Symbolic inductive reasoning, and System equation solving.

A key design decision is the integration with external, specialized tools (e.g., theorem provers, planning engines, symbolic algebra systems) to verify solutions in these highly expressive domains. This moves beyond simple answer-checking to assessing complex, structured outputs. Furthermore, Reasoning Core incorporates mechanisms for offline parallel generation and problem search, ensuring a continuous supply of novel and challenging instances. We provide a continuous "difficulty knob" for each generator, enabling adaptive curricula rather than relying on hardcoded levels. Our contribution is a new, open-ended playground for pushing LLMs towards more general and robust reasoning abilities. The code and data are publicly available\footnote{%
  \begin{tabular}{@{}ll@{}}
    \github & \url{https://github.com/sileod/reasoning_core} \\
    \hf & \url{https://hf.co/datasets/reasoning-core/rc1}\\
    \pitl & \href{https://app.primeintellect.ai/dashboard/environments/sileod/reasoning-core-env}{\texttt{environments/sileod/reasoning-core-env}}

  \end{tabular}%
}.

\section{Related Work}

\paragraph{Fixed Reasoning Benchmarks}
LLM reasoning research has been heavily driven by static benchmarks. These include mathematical problem-solving datasets like GSM8K~\cite{cobbe2021training} and MATH~\cite{hendrycksmath2021}, broad-spectrum reasoning collections like BIG-Bench~\cite{srivastava2022beyond} and GPQA~\cite{rein2024gpqa}, and code generation challenges~\cite{chen2021evaluatinglargelanguagemodels, jimenez2023swe}. The primary limitation of these benchmarks is their fixed nature, which makes them susceptible to dataset contamination and prevents the evaluation of generalization to novel problems of increasing difficulty~\cite{kiela2021dynabench, singh2025leaderboardillusion}.

\paragraph{Procedurally Generated Environments}
To overcome static dataset limitations, procedural content generation (PCG) has become popular for creating dynamic and scalable evaluation environments~\cite{cobbe2020leveraging, risi2020increasing}. Reasoning Gym~\cite{stojanovski2025reasoning} is a state-of-the-art example, offering a large suite of algorithmically verifiable tasks with parametric difficulty. Other works have focused on procedurally generating specific types of challenges, such as games~\cite{balrog, seely2025sudokubenchevaluatingcreativereasoning}, logic puzzles~\cite{lin2025zebralogicscalinglimitsllms, zhu2025autologiautomatedgenerationlogic}, or visual reasoning problems~\cite{chollet2025arc}. While crucial for robust evaluation, Reasoning Core differentiates itself by focusing not on games or toy problems like counting "legs" in an animal list, but on the full generality of fundamental symbolic domains like planning and formal logic, which are expressive enough to model a wide range of real-world problems.

\paragraph{RLVR and Synthetic Data}
The RLVR training paradigm, popularized by models like DeepSeek-R1~\cite{r1} and Tülu 3~\cite{tulu3}, relies on a continuous supply of problems with verifiable outcomes. This has spurred a line of research into creating large-scale synthetic datasets. Projects like SYNTHETIC-1~\cite{primeIntellectSynthetic1} and OpenThoughts~\cite{OpenThoughts} aim to generate millions of high-quality reasoning tasks. Other approaches propose self-evolutionary systems where a model generates its own curricula, using code executors for verification~\cite{zhao2025absolutezeroreinforcedselfplay}. Reasoning Core contributes to this ecosystem by providing a new source of high-quality, verifiable data that targets core, generalizable symbolic reasoning skills. By interfacing with external solvers, it ensures a rigorous standard of correctness for complex, symbolic domains that are difficult to verify otherwise.

\section{Reasoning Core}

\begin{figure*}[htbp]
\centering
\includegraphics[width=\textwidth]{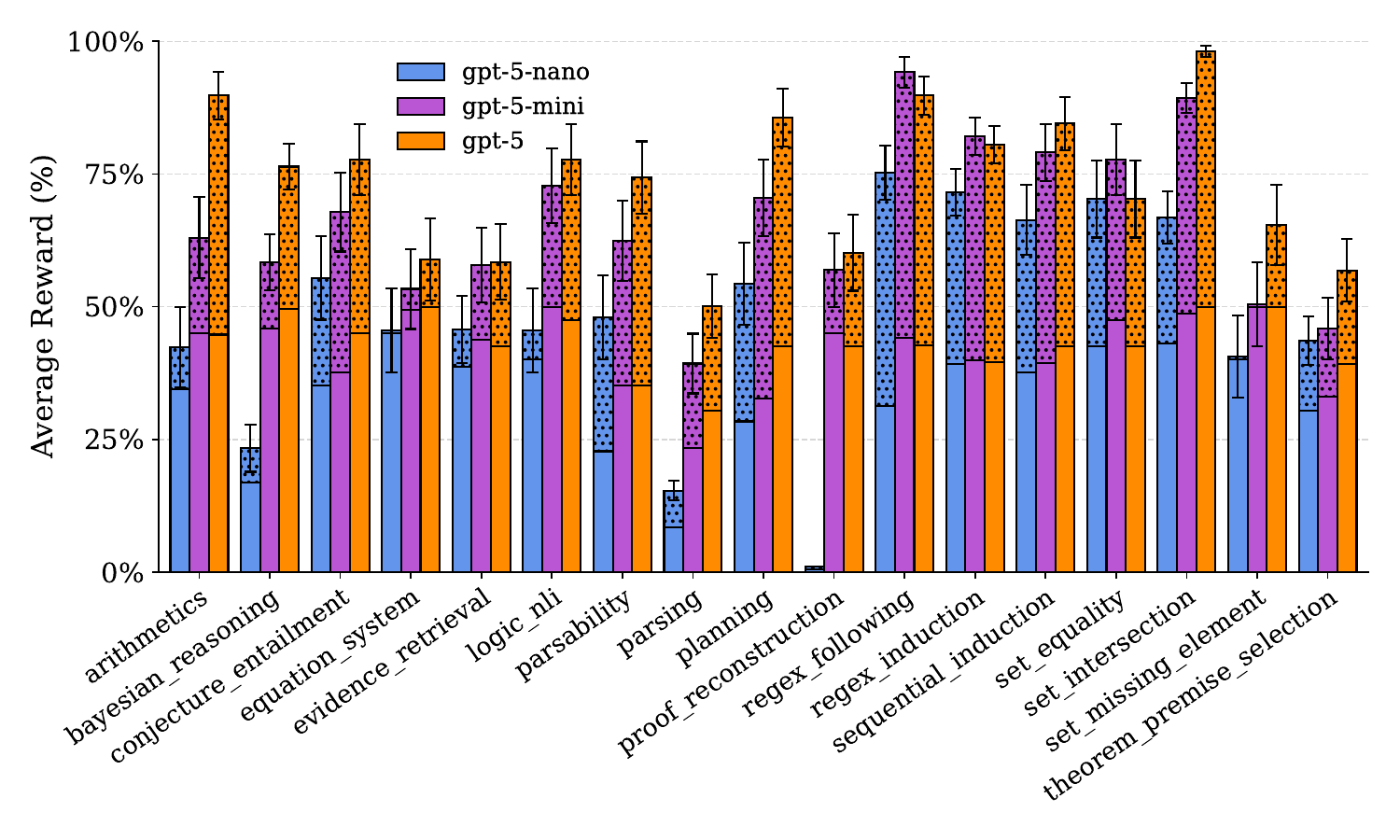}
\caption{Zero-shot average reward of GPT-5 on Reasoning Core tasks, evaluated across two difficulty levels. Solid bar ( \easySymbol\ )  is easy, dotted bar ( \hardSymbol\ ) is hard. Each bar represents the average reward for a task, indicating GPT-5's ability to solve problems without prior training on Reasoning Core data. The results demonstrate the challenging nature of the benchmark, particularly at higher difficulty settings.}
\label{fig:tasks}
\end{figure*}

We design Reasoning Core to address core symbolic reasoning challenges with a focus on generality, scalability, and verifiable rigor.

\paragraph{High Generality and Foundational Tasks}
Our task selection targets fundamental cognitive capacities rather than narrow puzzles. We include PDDL planning in randomly generated domains, not just fixed scenarios like Rubik's Cube. Formal logic tasks use full first-order logic with equality, allowing for deep inference rather than simple propositional assignments. We tackle system equation solving, covering non-linear systems, which probes algebraic manipulation and solution uniqueness. Additionally, context-free grammar parsing and regex induction assess fundamental linguistic and pattern-recognition skills. Symbolic induction and retrieval/perception tasks (e.g., missing element detection, set intersection) round out the suite, focusing on core symbolic learning and structured data manipulation. We provide causal reasoning tasks based on randomly sampled bayesian networks. We also include formal mathematics tasks by relying on the TPTP ecosystem, involving axiom/theorem matching, useful axiom detection, and proof structure tasks. These domains are chosen for their broad applicability and their demand for robust, transferable reasoning strategies.

Task descriptions are available in Appendix \ref{sec:task_desc}

\paragraph{Scalable Generation and Difficulty Control}
For each task, we implement a generator controlled by a continuous "difficulty knob", a single float value that parametrically adjusts hardcoded underlying factors (e.g., proof depth in logic, number of variables in equations, plan length in PDDL). For hyperparameters that are inherently discrete, we employ stochastic rounding based on the continuous value, allowing for fine-grained control over the generated problem distribution. This design facilitates the creation of adaptive curricula that can be tailored to a model's evolving performance, moving beyond rigid, predefined difficulty levels and ensuring a virtually infinite supply of novel training instances.

\paragraph{Verification via External Tools}
For the complex symbolic domains, internal verification alone is insufficient. We integrate external, specialized tools such as theorem provers for logic, planning engines for PDDL, and symbolic algebra systems for equation solving to provide objective and unambiguous reward signals. This ensures a high standard of correctness, crucial for robust RLVR training, and allows us to rigorously assess complex, structured outputs. These tools enable us to identify not only correct answers but also to handle nuances like problem solvability and solution optimality.

\paragraph{Grammar-based generation}
We use grammars where applicable to provide a concise and readable representation of data generators for arithmetic tasks, the regex following tasks, and the grammar understanding tasks (using a meta-grammar that produces grammars).
We use custom efficient scalable generation algorithms tailored to control not only the maximum but also the minimum depth of the generation to properly adjust the problems difficulty.

\paragraph{Efficient Data Production}
To support continuous RLVR training, Reasoning Core employs search-based and offline parallel generation pipelines. This architecture enables the rapid production of a vast number of diverse problems, ensuring that models are constantly exposed to novel challenges. This capability is critical for preventing overfitting and fostering robust, generalizable reasoning skills across a wide spectrum of problem instances.

\section{Results}

We conducted an initial zero-shot evaluation of GPT-5 with low reasoning effort on the Reasoning Core tasks, assessing its performance across two difficulty levels: easy (knob level 0) and hard (knob level 5), with 200 samples per task and difficulty. Figure \ref{fig:tasks} presents the average reward obtained for each task configuration. All tasks are sufficiently challenging for GPT-5, and the difficulty control works as intended for most of the tasks, leading to higher failure rates with the hard mode.

\section{Conclusion}
We introduced Reasoning Core, a library of scalable RL environments targeting fundamental symbolic reasoning capacities in LLMs. By focusing on high-generality tasks like PDDL planning and first-order logic, and leveraging external solvers for rigorous verification, we aim to overcome the limitations of existing fixed and procedurally generated benchmarks. Our framework's continuous difficulty control and efficient data generation provide a powerful tool for developing robust and transferable reasoning skills in LLMs. The initial zero-shot performance of GPT-5 underscores the demanding nature of Reasoning Core, positioning it as an important resource for future RLVR research.

\bibliographystyle{acl_natbib}
\bibliography{custom}
\onecolumn
\appendix
\section{Task descriptions \label{sec:task_desc}}

\subsection{\texttt{planning}}

This task challenges models to generate valid action sequences in procedurally generated planning domains. Each instance presents a randomly constructed PDDL-like domain with objects, actions, preconditions, and effects, where models must reason about state transitions to achieve specified goals. The generator creates diverse planning scenarios by varying domain complexity, object types, fluent arities, and goal configurations, ensuring broad coverage of planning reasoning patterns. Solutions are validated for both syntactic correctness and semantic validity against the domain constraints.
Unlike previous generators of planning data, this is the first, to our knowledge, to use randomly sample domains instead of using existing ones (e.g., BlocksWorld or Sokoban);

\begin{promptanswer}
\begin{PAprompt}
I am playing with a set of objects.

Here are the actions I can do:
action_0 with action_0_parameter0

I have the following restrictions on my actions:

Once action_0 action is performed the following facts will be true: fluent_0(action_0_parameter0).

Everything unspecified is false by default

My goal is to have that fluent_0(object_1), fluent_0(object_2).
Hint: Reference solution has 2 actions (may not be optimal). Return only the plan:
Multiple lines if needed, one action i.e. actionx(objectx, objectx...) per line.
\end{PAprompt}

\begin{PAanswer}
    action_0(object_2)
    action_0(object_1)
\end{PAanswer}
\end{promptanswer}

\subsection{\texttt{equation\_system}}

This task evaluates the ability to solve systems of linear equations and correctly identify when systems are underdetermined or inconsistent. The generator constructs systems by starting with a known unique solution and applying random linear combinations to obfuscate the equations, then probabilistically modifies the system to create inconsistent cases (by adding contradictory constraints) or underdetermined cases (by removing equations). The distribution spans various system configurations including well-posed problems with unique solutions, overconstrained inconsistent systems, and underconstrained systems with multiple solutions.

\begin{promptanswer}
\begin{PAprompt}
Solve the following system of equations for the variable 'X2'.

System:
  X1 + 13 = 0
  X3 - 21 = 0

Return the numerical value for X2. If a unique numerical solution does not exist, return either 'No solution' or 'Multiple solutions'.
\end{PAprompt}

\begin{PAanswer}
Multiple solutions
\end{PAanswer}
\end{promptanswer}

\subsection{\texttt{regex\_following}}

This task evaluates the ability to generate valid strings that match given regular expressions. The model is presented with a regex pattern and must produce a string that would be accepted by that pattern when evaluated with full-match semantics. The regex patterns are generated using a context-free grammar that produces expressions of varying complexity, incorporating character classes, quantifiers, alternation, grouping, and predefined character sets. The task includes both simple patterns involving literal characters and complex nested structures with multiple operators, providing a comprehensive test of regex comprehension across different syntactic constructs.

\begin{promptanswer}
\begin{PAprompt}
'daf' is a valid match for regex '[a-z]{3}' but not 'ab1'
Return a valid match for \.?T?+
\end{PAprompt}

\begin{PAanswer}
.
\end{PAanswer}
\end{promptanswer}

\subsection{\texttt{regex\_induction}}

This task challenges models to induce regular expressions from positive and negative examples. The task generator first creates a target regex using a probabilistic context-free grammar that combines character classes, quantifiers, alternation, concatenation, and predefined patterns like \texttt{\textbackslash d} and \texttt{\textbackslash w}. For each problem instance, positive examples are generated by sampling strings that match the target regex, while negative examples are created by sampling from different randomly generated regexes and filtering out any that accidentally match the target. The scoring function rewards solutions that achieve perfect classification accuracy on the provided examples, with a length penalty applied only when accuracy reaches 100\% to prefer more concise expressions.

\begin{promptanswer}
\begin{PAprompt}
Return a regex that matches all POSITIVE strings and none of the NEGATIVE strings.
POSITIVE: 'Q\X', 'G', 'IU', 'UG', 'RUYU', 'Y', 'H', 'BUE'
7^', 'n8', '	', '7Ts', 'o', 'Í9.', '.'
\end{PAprompt}

\begin{PAanswer}
(([9-a])+?)
\end{PAanswer}
\end{promptanswer}

\subsection{\texttt{arithmetics}}

This task generates arithmetic expressions with diverse structural complexity and numerical configurations. Expressions are constructed using a context-free grammar that recursively combines basic operations (addition, subtraction, multiplication, division, squaring) and parenthetical groupings with randomly sampled numerical values. The generator ensures mathematical validity by rejecting divisions by zero and constraining output precision, while maintaining broad coverage through configurable expression depth, decimal precision, and floating-point probability parameters. Models must evaluate the complete expression and provide the numerical result, testing computational reasoning across varying levels of syntactic and arithmetic complexity.

\begin{promptanswer}
\begin{PAprompt}
Evaluate (-12 + 13 * 10) * 7.6 - 8.1 * 5.1 + (-2.6) + 12 + (5.6 / -14).
 Answer with only a number.
\end{PAprompt}

\begin{PAanswer}
864.49
\end{PAanswer}
\end{promptanswer}

\subsection{\texttt{sequential\_induction}}

This task challenges models to infer recursive formulas from numerical sequences. Given a sequence of numbers and its degree of recursion, models must deduce the underlying mathematical relationship that generates subsequent terms based on previous values and the current index. The task generates sequences using context-free grammars that produce recursive formulas with configurable complexity, ranging from simple arithmetic operations to more sophisticated functions including modular arithmetic and unary operators. The evaluation framework ensures mathematical validity by filtering out degenerate cases and controlling for numerical explosion, creating a robust testbed for sequence reasoning and formula induction capabilities.

\begin{promptanswer}
\begin{PAprompt}
You are given a sequence of 8 numbers generated by a recursive formula of known degree of recursion, here equal to 1.
The indexation of the sequence start from 0, i.e. we provide you [U0,U1,...,U7].Your task is to infer the formula that defines U[n] in terms of previous values and the current index n.

Instruction:
- Use only the binary operators: +, -, *, **
- Reference if necessary to previous terms as U[n - 1], U[n - 2], ..., U[n - d] where d is the given degree of recursion (use exactly this format)
- You can use "n" as the current index (e.g., U[n] = n)
- You must only provide the right-hand side of the formula (i.e., f(U[n], n) such that U[n] = f(...))
- ⚠️ This implies to not include "U[n] =" in your output.
 - The degree of recursion of your guessed formula must be inferior or equal to the one of the true formula

- The sequence you are asked to induce its recursive formula have the following properties:
Sequence: [-1, -1, -2, -6, -24, -120, -720, -5040]
Degree of recurrence: 1
Initial terms: [-1]

 Your provided answer must be valid for all terms n ≥ d, and must be as simple as possible.
\end{PAprompt}

\begin{PAanswer}
n*U[n - 1]
\end{PAanswer}
\end{promptanswer}

\subsection{\texttt{conjecture\_entailment}}

This task generates problems following \citep{quesnel2025saturation} to determine whether a given subset of axioms is sufficient to prove a specific theorem. The generator constructs derivation graphs from TPTP axiom sets across diverse mathematical domains including geometry, algebra, set theory, and topology, then extracts theorems at configurable proof depths. For each problem, the system either provides the correct minimal axiom subset (positive case) or introduces perturbations through axiom replacement, addition, or removal to create unprovable instances, with theorem provability verified using the Vampire prover and superposition calculus.

\begin{promptanswer}
\begin{PAprompt}
You are a mathematical logic assistant. Your task is to determine the sufficiency of a specific set of axioms for proving a theorem.

By using the **Superposition Calculus** (which includes rules like Resolution and Paramodulation).
## General Context
The problem is set in the domain of: **Geometry**.
The following are the fundamental axioms of this domain, providing a general theoretical background:
Fundamental Axioms:
- cnf(transitivity_for_equidistance,axiom,(equidistant(X3,X4,X5,X6)|~equidistant(X1,X2,X3,X4)|~equidistant(X1,X2,X5,X6)))
- cnf(reflexivity_for_equidistance,axiom,(equidistant(X1,X2,X2,X1)))

--- 

## Task
Now, you are given a specific **subset of axioms** and a theorem from this domain.

**Axiom Subset under consideration:**
- (equidistant(X1,X2,X3,X4)|~equidistant(X4,X3,X1,X2))
- (equidistant(X1,X2,X3,X4)|~equidistant(X5,X6,X1,X2)|~equidistant(X4,X3,X5,X6))

**Theorem to prove:**
`(equidistant(X1,X2,X3,X4)|~equidistant(X4,X3,X5,X6)|~equidistant(X2,X1,X5,X6))`

### Question
Is the **"Axiom Subset under consideration"** listed above **sufficient on its own** to prove the **"Theorem to prove"**?

### Response Format
Respond **only** with `True` if the provided subset is sufficient, or `False` otherwise. Do not provide explanations.
\end{PAprompt}

\begin{PAanswer}
True
\end{PAanswer}
\end{promptanswer}

\subsection{\texttt{theorem\_premise\_selection}}

This task requires identifying the minimal subset of premises necessary to prove a given theorem within various mathematical domains including algebra, geometry, group theory, and ring theory. The system generates problems by extracting theorems from automatically constructed derivation graphs using the Superposition Calculus, then creates pools of candidate premises that include both necessary premises and distractors. Models must select only those premises that are both necessary and sufficient for the proof, testing their ability to distinguish essential logical dependencies from irrelevant information across diverse mathematical contexts.

\begin{promptanswer}
\begin{PAprompt}
You are a mathematical logic assistant. Your task is to identify a minimal set of premises sufficient for a proof.

By using the **Superposition Calculus** (which includes rules like Resolution and Paramodulation).
## General Context
The problem is set in the domain of: **Ring Theory**.
The following are the fundamental axioms of this domain. They provide general context. **Do not use them in the proof itself.**
Fundamental Axioms:
- cnf(distribute1,axiom,(multiply(X1,add(X2,X3))=add(multiply(X1,X2),multiply(X1,X3))))
- cnf(distribute2,axiom,(multiply(add(X1,X2),X3)=add(multiply(X1,X3),multiply(X2,X3))))

--- 

## Task
Your goal is to prove the following theorem:
**Theorem:**
`(multiply(add(X1,X1),X2)=multiply(X1,add(X2,X2)))`

Below is a numbered pool of potential premises. Your task is to identify the **minimal subset** of numbers from this pool whose corresponding statements are **sufficient on their own** to prove the theorem.
**Pool of Premises:**
1. (multiply(X1,add(X2,X3))=add(multiply(X1,X2),multiply(X1,X3)))
2. (add(multiply(X1,multiply(X2,X3)),multiply(X4,multiply(X2,additive_inverse(X3))))=multiply(add(X1,additive_inverse(X4)),multiply(X2,X3)))
3. (add(multiply(X1,add(X2,X2)),add(X3,multiply(add(X1,X1),additive_inverse(X2))))=X3)
4. (multiply(add(X1,X2),X3)=add(multiply(X1,X3),multiply(X2,X3)))

### Question
Which is the smallest set of numbered premises from the pool that is sufficient to prove the theorem, without using the fundamental axioms from the context?

### Response Format
Your answer must be **only** a list of numbers, sorted in increasing order. For example: `[2, 5, 8]`.
\end{PAprompt}

\begin{PAanswer}
[1, 4]
\end{PAanswer}
\end{promptanswer}

\subsection{\texttt{proof\_reconstruction}}

This task challenges models to reconstruct logical proof dependency graphs from shuffled mathematical clauses. Given a theorem and its proof components presented in random order, models must identify which clauses serve as axioms and determine the exact derivation relationships between derived clauses and their parent premises. The task operates across diverse mathematical domains including algebra, analysis, geometry, and set theory, with proofs generated using the Superposition Calculus ensuring that each derived clause follows from exactly two parent clauses. Evaluation combines structural correctness of the reconstructed proof graph with semantic validation of individual derivation steps using automated theorem provers.

\begin{promptanswer}
\begin{PAprompt}
Your task is to reconstruct the dependency graph of a mathematical proof from the domain of **Analysis**.

The proof graph concludes with the theorem: `(minimum(X1,X1)=X1)`

## Proof Context & Rules
This proof was generated by using the **Superposition Calculus** (which includes rules like Resolution and Paramodulation).

Therefore, the proof has the following properties:
- **Starting Points:** Some clauses in the list are starting points (axioms ) and are not derived from other clauses.
- **Derived Clauses:** Every other clause is derived from exactly **two** parent clauses from the list.
- **Clause Reuse:** A single clause can be used as a parent in multiple derivation steps.

## Your Task
Given the rules above, reconstruct the proof from the following shuffled list of clauses. Identify the derivation for every clause that is not a starting point.

**Shuffled Clauses:**
1. (minimum(X1,X1)=X1)
2. (minimum(X2,X1)=X1|~less_or_equal(X1,X2))
3. (less_or_equal(X1,X1))

## Required Output Format
- List **only** the derivation steps.
- Each step must be on a new line.
- Use the exact format `CHILD <- PARENT_1, PARENT_2`. Example: `5 <- 2, 4`.(for each line)
- All clauses from the list must be used in the final structure.
- No explanations, comments, or extra text.
\end{PAprompt}

\begin{PAanswer}
1 <- 2, 3
\end{PAanswer}
\end{promptanswer}

\subsection{\texttt{logic\_nli}}

This task generates natural language inference problems grounded in first-order logic, following \citep{sileo-2024-scaling}. The generator creates premises consisting of multiple logical statements about entities in a room, along with hypotheses that may be entailed, contradicted, or neither by the premises. Automated theorem proving determines the correct logical relationship, ensuring training examples span diverse reasoning patterns including universal quantification, existential statements, and complex predicate relationships. The task provides formal logical proofs as metadata, enabling analysis of reasoning steps required for each inference.

\begin{promptanswer}
\begin{PAprompt}
Premise:
there is a room.
all scarred persons in the room are humble
everyone in the room owns a 3D printer if they develops open-source software projects in their free time
everyone in the room is long haired
someone in the room is a quiet patient person
no wise person in the room is creative
Lucy is a funny person
Susan neither is quiet nor is an active member of a local robotics club
Hypothesis:
Lucy is funny

If the Premise entails the Hypothesis, the label is 'entailment'.
If the Premise contradicts the Hypothesis, the label is 'contradiction'.
If neither, the label is 'neutral'.
Answer with exactly one word, neutral|contradiction|entailment
\end{PAprompt}

\begin{PAanswer}
entailment
\end{PAanswer}
\end{promptanswer}

\subsection{\texttt{evidence\_retrieval}}

This task requires models to identify which specific statements within a logical premise support entailment or contradiction relationships with a given hypothesis, following \citep{sileo2025logic}. The task generates diverse first-order logic problems with premises containing multiple statements and asks models to select only those statements that are necessary for proving the logical relationship. Generated problems span varying numbers of premises and logical complexity levels, ensuring models must perform precise logical reasoning rather than surface-level pattern matching. The training instances are automatically verified using theorem provers to guarantee correctness of the required evidence sets.

\begin{promptanswer}
\begin{PAprompt}
Premise:
[0] Mary, Paul, Fred are the only persons in the room.
[1] Mary can play the flute
[2] everyone in the room who does not travel domestically frequently neither is a cybersecurity expert nor practices archery
[3] if someone is a funny popular person then he/she is a cybersecurity expert
[4] everyone in the room is organized, is not a quiet person and can play the flute
[5] John is humble
[6] everyone in the room who enjoys windsurfing is a funny person
[7] Alice is a curious person
Hypothesis:
Paul is a quiet person

Which statements in the premise contradict the hypothesis?
Only answer the list of supporting statements, e.g. [0, 6, 7].
\end{PAprompt}

\begin{PAanswer}
[0, 4]
\end{PAanswer}
\end{promptanswer}

\subsection{\texttt{parsability}}

This task evaluates the ability to determine whether a given string can be parsed by a context-free grammar, and if so, whether the parse is unique or ambiguous. The generator creates diverse context-free grammars with varying numbers of nonterminals and terminals, incorporating structural elements like nested expressions and bracket pairs. For each grammar, strings are generated either through valid derivations or through perturbations that may introduce unparsable sequences. The task requires models to analyze the relationship between formal grammars and strings, classifying each string-grammar pair as unambiguous, ambiguous, or unparsable.

\begin{promptanswer}
\begin{PAprompt}
(GRAMMAR)
S -> B
    B -> '[' 'of' 'expert' ']' 'development'
    C -> C 'seek'
    B -> B
    C -> 'your' A A

(STRING)
of [ expert development ]

(QUESTION)
What is the parsability of this string?
Answer with exactly one word, unambiguous|ambiguous|unparsable
\end{PAprompt}

\begin{PAanswer}
unparsable
\end{PAanswer}
\end{promptanswer}

\subsection{\texttt{parsing}}

This task requires models to generate fully parenthesized parse trees in Lisp-style notation for strings given context-free grammars. The task employs a meta-grammar approach to generate diverse CFGs with varying numbers of nonterminals, terminals, and production rules, including nested structures with Dyck languages. Each instance provides a grammar specification and an input string that is guaranteed to have exactly one valid parse tree, and models must produce the complete syntactic analysis with proper bracketing and case conventions (uppercase nonterminals, lowercase terminals). The task evaluates both syntactic parsing competence and the ability to follow structured output formatting requirements across a broad distribution of grammatical forms.

\begin{promptanswer}
\begin{PAprompt}
(GRAMMAR)
S -> B
    D -> 'shake' 'reach'
    B -> 'shake' B
    B -> 'reach'

(STRING)
shake shake shake shake reach

(QUESTION)
Return the fully parenthesized parse tree of STRING in Lisp style.
Use uppercase for nonterminals, lowercase unquoted tokens for terminals
Given G_ex: S -> NP VP, NP -> 'det' Noun, Noun -> 'noun', VP -> 'verb'         and G_ex: "det noun verb" correct Lisp Parse Tree would be (S (NP det (Noun noun)) (VP verb))."
        
\end{PAprompt}

\begin{PAanswer}
(S (B shake (B shake (B shake (B shake (B reach))))))
\end{PAanswer}
\end{promptanswer}

\subsection{\texttt{bayesian\_association}}

This task evaluates probabilistic reasoning within Bayesian networks by requiring models to compute posterior probability distributions given observational evidence. The task generates random directed acyclic graphs with specified numbers of variables and domain sizes, where each variable follows conditional probability distributions dependent on its parents in the network. Models must perform exact Bayesian inference to calculate the probability distribution of a target variable conditioned on observed values of other variables in the system. The task assesses fundamental skills in probabilistic reasoning and conditional independence relationships that are essential for understanding causal structures and making predictions under uncertainty.

\begin{promptanswer}
\begin{PAprompt}
### System Description
This section describes the probabilistic relationships between variables in the system:
the probability of X00 = 0 is 0.7 and the probability of X00 = 1 is 0.3. 
 If X00 = 0, then the probability of X01 = 0 is 0.34 and the probability of X01 = 1 is 0.66. 
 If X00 = 1, then the probability of X01 = 0 is 0.35 and the probability of X01 = 1 is 0.65. 
 If X00 = 0, then the probability of X02 = 0 is 0.08 and the probability of X02 = 1 is 0.92. 
 If X00 = 1, then the probability of X02 = 0 is 0.65 and the probability of X02 = 1 is 0.35.

### Scenario
Given the system described above, consider the following specific conditions:
Observing/Knowing that the state X02 is equal to 0

### Your Task
Calculate the probability distribution for the variable 'X00', which can take the following values: ['0', '1'].

### Required Output Format
You must return the probability distribution over all values of the target variable in the format of a Python dictionary. The output should map each value to its estimated probability.
You will be evaluated based on how close your estimated probability distribution is to the true one.

For example, if the target variable is X01 (which can take values 0 or 1) and you estimate that P(X01 = 0) = 0.4 and P(X01 = 1) = 0.6, your answer must be: {0: 0.4, 1: 0.6} (in between the proper xml tags if asked). 
\end{PAprompt}

\begin{PAanswer}
{0: 0.2267360798241549, 1: 0.7732639201758451}
\end{PAanswer}
\end{promptanswer}

\subsection{\texttt{bayesian\_intervention}}

This task evaluates causal reasoning capabilities by requiring models to compute probability distributions under interventions in Bayesian networks. Given a system of probabilistic relationships between variables described in natural language, models must calculate the posterior distribution of a target variable when both performing an intervention (using the do-operator) on one variable and observing evidence about others. The task generates random directed acyclic graphs with configurable numbers of nodes and domain sizes, then applies do-calculus to determine the correct interventional distributions, testing the model's ability to distinguish between observational and interventional queries in causal inference.

\begin{promptanswer}
\begin{PAprompt}
### System Description
This section describes the probabilistic relationships between variables in the system:
the probability of X00 = 0 is 0.78 and the probability of X00 = 1 is 0.22. 
 If X00 = 0, then the probability of X01 = 0 is 0.59 and the probability of X01 = 1 is 0.41. 
 If X00 = 1, then the probability of X01 = 0 is 0.5 and the probability of X01 = 1 is 0.5. 
 If X01 = 0, then the probability of X02 = 0 is 0.78 and the probability of X02 = 1 is 0.22. 
 If X01 = 1, then the probability of X02 = 0 is 0.01 and the probability of X02 = 1 is 0.99.

### Scenario
Given the system described above, consider the following specific conditions:
Doing/Imposing that the state X02 is equal to 0. Observing/Knowing that the state X00 is equal to 1

### Your Task
Calculate the probability distribution for the variable 'X01', which can take the following values: ['0', '1'].

### Required Output Format
You must return the probability distribution over all values of the target variable in the format of a Python dictionary. The output should map each value to its estimated probability.
You will be evaluated based on how close your estimated probability distribution is to the true one.

For example, if the target variable is X01 (which can take values 0 or 1) and you estimate that P(X01 = 0) = 0.4 and P(X01 = 1) = 0.6, your answer must be: {0: 0.4, 1: 0.6} (in between the proper xml tags if asked). 
\end{PAprompt}

\begin{PAanswer}
{0: 0.5015292832281123, 1: 0.4984707167718876}
\end{PAanswer}
\end{promptanswer}

\subsection{\texttt{set\_equality}}

This task evaluates set equality reasoning by presenting two lists and requiring the model to determine whether they contain exactly the same elements regardless of order. The generator creates a base set by randomly sampling from diverse domains including integers, spelled-out numbers in multiple languages, dates in various formats, and letter combinations. With configurable probability, the second set is either an exact shuffle of the first (yielding True) or a perturbed version where elements are randomly added, removed, or replaced (yielding False). The perturbation operations and set sizes scale with difficulty level, providing systematic training for fundamental set comparison skills across heterogeneous data types.

\begin{promptanswer}
\begin{PAprompt}
Set1: [284, 486, 734, 380, 933, 874, 152, 348, 169, 898]
Set2: [380, 169, 898, 152, 486, 284, 933, 348, 874, 734]
Only return True if Set1 and Set2 contain exactly the same elements, False otherwise.
\end{PAprompt}

\begin{PAanswer}
True
\end{PAanswer}
\end{promptanswer}

\subsection{\texttt{set\_intersection}}

This task evaluates the ability to compute set intersections by presenting two sets and requiring the model to identify their common elements. The generator creates pairs of sets where the first set is randomly sampled from a chosen domain, and the second set consists of elements both inside and outside the first set to ensure non-trivial intersections. The task operates across diverse data types including integers, multilingual number words (English and French), dates in multiple formats, and automatically generated letter strings, providing robust coverage of symbolic reasoning scenarios. Answers are evaluated using Jaccard similarity to allow partial credit for approximately correct responses.

\begin{promptanswer}
\begin{PAprompt}
Set1: [912, 986, 769, 838, 833, 34, 76, 166, 786, 560]
Set2: [34, 327, 846, 135, 166, 838]
Only return the intersection of Set1 and Set2 as a Python set: {elem_1, elem_2, ..., elem_n}.
\end{PAprompt}

\begin{PAanswer}
{34, 166, 838}
\end{PAanswer}
\end{promptanswer}

\subsection{\texttt{set\_missing\_element}}

This task evaluates the ability to identify missing elements in contiguous sequences, following \citep{sileo2024attention}. The generator creates an ordered subsequence from diverse domains including integers, natural language numbers (English/French), dates, and letter combinations, then removes one internal element and presents the remaining elements in shuffled order. Models must infer the underlying sequential structure and determine which element has been systematically omitted. The task spans multiple semantic domains to assess generalization across different representational formats and sequential patterns.

\begin{promptanswer}
\begin{PAprompt}
Set_A: ['bz', 'by', 'ca', 'bw', 'bt', 'cb', 'bv', 'bs', 'bu']
Only return the string element missing from Set_A.
\end{PAprompt}

\begin{PAanswer}
bx
\end{PAanswer}
\end{promptanswer}

\end{document}